\documentclass{egpubl}
\usepackage{eurovis2025}

\SpecialIssuePaper         

\CGFccby

\usepackage[T1]{fontenc}
\usepackage{dfadobe}  

\usepackage{cite}  
\BibtexOrBiblatex
\electronicVersion
\PrintedOrElectronic
\ifpdf \usepackage[pdftex]{graphicx} \pdfcompresslevel=9
\else \usepackage[dvips]{graphicx} \fi

\usepackage{egweblnk}

\usepackage{amsmath}
\usepackage{amssymb}
\usepackage{subcaption}
\usepackage{nicefrac}

\title[Dimensionality Reduction Meets Graph (Drawing) Theory]%
      {When Dimensionality Reduction Meets Graph (Drawing) Theory: Introducing a Common Framework, Challenges and Opportunities\vspace{-1cm}}

\author[Paulovich et al.]
{\parbox{\textwidth}{\centering  
F. V. Paulovich$^{1}$\orcid{0000-0002-2316-760X}, A. Arleo$^{1}$\orcid{0000-0003-2008-3651}, and S. van den Elzen$^{1}$\orcid{0000-0003-1245-0503}}
        \\
{\parbox{\textwidth}{\centering $^1$Eindhoven University of Technology (TU/e), The Netherlands
      }
}
}

\begin{document}


\teaser{
    \vspace{-1cm}
    \centering
    \includegraphics[width=0.85\linewidth]{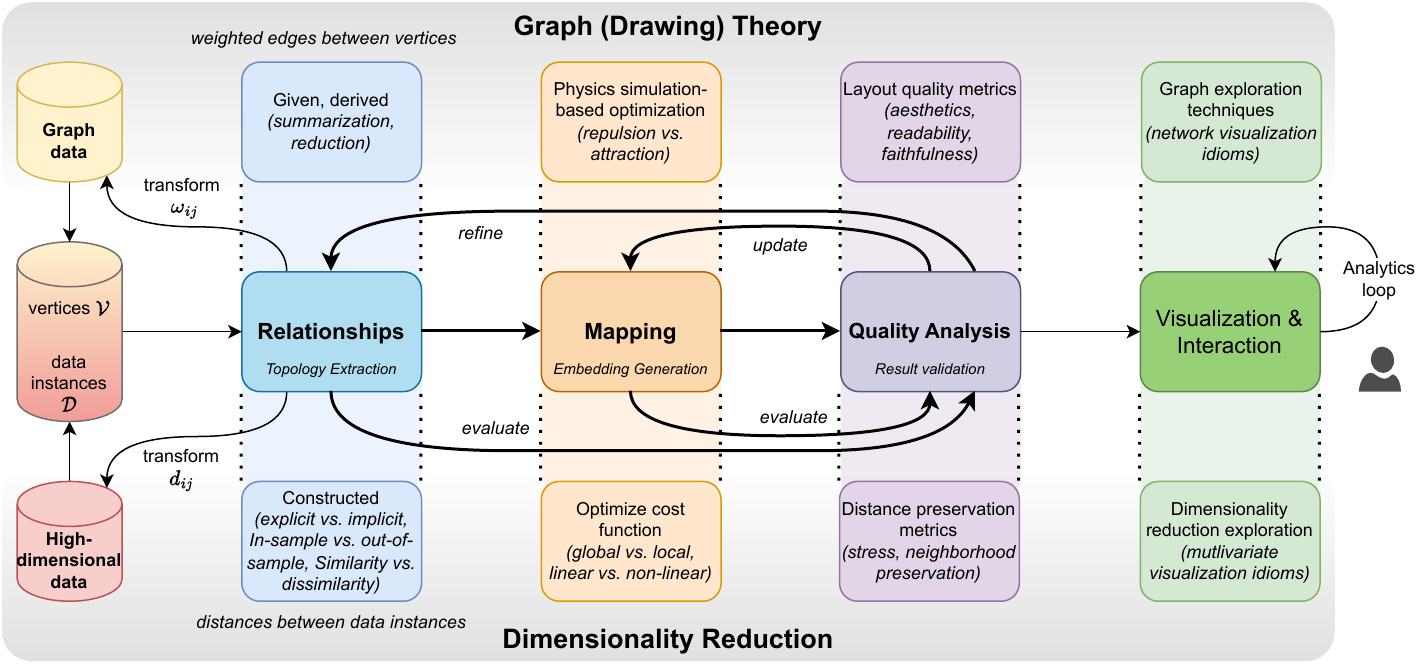}
    \caption{Our approach explores the integration and alignment of the various stages in Dimensionality Reduction (DR) and Graph Drawing, demonstrating that strategies and algorithms in both areas share similar objectives but use complementary methods. With topology extraction techniques, high-dimensional data can be converted into graph data and vice versa. In the context of DR, distances are defined based on facets of the data, creating a matrix that simulates a weighted, fully-connected graph. The relationships facilitate the creation of a mapping, with subsequent quality analysis comparing the two, and enabling iterative updates. The outcome of this mapping is visualized, offering interaction for human-in-the-loop analytics. Linking concepts and stages from both domains creates substantial opportunities for cross-fertilization.}
    \label{fig:framework}
}

\maketitle
\begin{abstract}
   In the vast landscape of visualization research, Dimensionality Reduction (DR) and graph analysis are two popular subfields, often essential to most visual data analytics setups. DR aims to create representations to support neighborhood and similarity analysis on complex, large datasets. Graph analysis focuses on identifying the salient topological properties and key actors within networked data, with specialized research on investigating how such features could be presented to the user to ease the comprehension of the underlying structure. Although these two disciplines are typically regarded as disjoint subfields, we argue that both fields share strong similarities and synergies that can potentially benefit both. Therefore, this paper discusses and introduces a unifying framework to help bridge the gap between DR and graph (drawing) theory. Our goal is to use the strongly math-grounded graph theory to improve the overall process of creating DR visual representations. We propose how to break the DR process into well-defined stages, discussing how to match some of the DR state-of-the-art techniques to this framework and presenting ideas on how graph drawing, topology features, and some popular algorithms and strategies used in graph analysis can be employed to improve DR topology extraction, embedding generation, and result validation. We also discuss the challenges and identify opportunities for implementing and using our framework, opening directions for future visualization research.


\end{abstract}  

\section{Introduction}\label{se:intro}

Dimensionality Reduction (DR) is commonly described by the visualization community as the process of mapping data items from a high-dimensional space to points in a lower dimensional visual space, usually 2D or 3D~\cite{8851280, 8383983}. The resulting visual representations aim to capture some trait of the data structure (either a local or global structure). On the other hand, Graph Drawing (GD) is the discipline rooted in graph theory and information visualization that investigates how to construct readable geometric visual representations of graphs in the 2D plane such that the geometry of the drawing faithfully resembles the topology of the original data structure~\cite{di2017graph}. 

Although somewhat separate disciplines, GD has already benefited from strategies introduced by DR methods. For instance, \textit{DRGraph}~\cite{minfeng2021drgraph} accelerates the embedding of large graphs using a multi-level methodology based on t-SNE~\cite{van2008visualizing}, borrowing ideas early introduced by Walshaw~\cite{Walshaw_2003} to improve running times and drawing quality. The opposite perspective, benefiting DR from GD has also been implicitly and explicitly discussed, among others, in the t-SNE~\cite{van2008visualizing} and UMAP papers~\cite{mcinnes2018umap}. Drawing methods, however, are not the only algorithms in the intersection of DR and GD. Topology transformation has also been considered. In a recent paper, Bot et al.~\cite{bot2024manifold}, introduce a minimum-spanning-tree-based manifold approximation capable of crossing sampling gaps without introducing shortcuts, increasing robustness of DR algorithms in case of non-uniformly sampled manifolds. 

These examples suggest a strong interconnection between the two disciplines that goes beyond just the calculation of the embedding. Our claim is that GD and DR methods are not alternatives but rather complementary strategies and that their interplay, although already considered, remains largely unexplored. In this paper, our focus is to better formalize the interactions between DR and GD, focusing on leveraging the opportunities concerning the integration of GD theory and techniques within the DR visualization pipeline. We use the visualization reference model as a theoretical basis on which to ground our work~\cite{mazza2009introduction}. This model describes the generation of a visual representation of data as a multi-stage process: (i) data preparation, including processing and transformations; (ii) visual mapping and view creation; and (iii) insight generation through analysis and (iv) exploration. Given the networked nature of the data at hand, we rename these stages as follows: \textbf{Relationships} (or \textit{topology extraction}), \textbf{Mapping} (or \textit{embedding generation}), \textbf{Quality Analysis} (or \textit{result validation}), and \textbf{Visualization \& Interaction} (see Figure~\ref{fig:framework}). Within this context, we outline how GD theory and drawing techniques can contribute to the efficacy of DR approaches in generating, validating, and visualizing an embedding of a DR layout in the plane.

Our contributions are as follows. We introduce and formalize a framework to map the interplay between DR and GD. Within this framework, we outline both acknowledged methodologies and new opportunities from graph theory to integrate within the DR pipeline, with potential new ideas for further work. Our goal is to raise awareness of unexploited opportunities at the intersection of DR, GD, and network visualization, defining a common terminology to foment opportunities to exploit the synergies between both fields.

 %


\section{Related Work and Background}
\label{sec:related}

In this section, the literature related to our solution is discussed. It is not our intention to survey the DR and GD field to map it within our proposed framework, but rather to present the important terminology and concepts that delineate our approach. For an extensive and detailed discussion of technical aspects, applications, and comparison of DR techniques, we suggest~\cite{van2009dimensionality, 8383983, 8851280}. For GD and network visualization, we refer the interested reader to the meta-survey by Filipov et al.~\cite{filipov2023survey}. 

Dimensionality Reduction (DR) has multiple definitions. 
In mathematical terms, let's ${\mathcal{X}=\{x_i, \ldots, x_N\} \in \mathbb{R}^m}$ be a set of $N$ data items represented in an $m$-dimensional space and ${\mathcal{Y}=\{y_i, \ldots, y_N\} \in \mathbb{R}^{n=\{2,3\}}}$ their mappings to the visual space. A DR technique can be viewed as a function ${f:\mathcal{X} \rightarrow \mathcal{Y}}$, which maps each data item $x_i$ into a point $y_i$, seeking to preserve relationships between items defined on the high-dimensional space as much as possible on the mapped points~\cite{8851280}.

The existing literature on DR offers different taxonomies and categorizations of the existing techniques. A common classification splits the techniques into two major groups: global and local techniques~\cite{mcinnes2018umap}. While in global techniques, full pairwise relationships among data items are considered, in local, only neighborhoods are used to represent how data items are related, for instance, how data items are similar or dissimilar. Examples of global techniques involve classical Multidimensional Scaling (MDS)~\cite{torgerson1952multidimensional}, Random Projection (RP)~\cite{ACHLIOPTAS2003671}, Part-linear Multidimensional Projections (PLMP)~\cite{5613468}, Force Scheme (FS)~\cite{doi:10.1057/palgrave.ivs.9500054} or IDMAP~\cite{10.1117/12.650880}, LAMP~\cite{6065024}, and Sammon's Mapping (SM)~\cite{sammon1969nonlinear}. Examples of local techniques involve Stochastic Neighborhood Embedding (SNE)~\cite{NIPS2002_6150ccc6}, t-Student Stochastic Neighborhood Embedding (t-SNE)~\cite{van2008visualizing}, UMAP~\cite{mcinnes2018umap}, ISOMAP~\cite{tenenbaum2000global}, LLE~\cite{roweis2000nonlinear}, LSP~\cite{4378370}, and LoCH~\cite{FADEL2015546}. 

Considering the nature of the embedding function $f$, another categorization splits DR techniques into linear or non-linear transformations. Linear functions can be viewed as transformations between coordinate vector spaces, such as PCA, MDS, PLMP, and Random Projection. They are easier to understand since the connection between the visual layout spatialization and the original feature $m$-dimensional space can be established (e.g., PCA biplots). The downside is that they cannot capture the data structure sampled over complex manifolds~\cite{8851280}. In those cases, non-linear techniques are preferred, such as t-SNE, IDMAP, UMAP, and Sammon's Mapping. However, understanding the produced layouts and controlling the involved parametrization can be challenging, usually presenting a stochastic non-deterministic nature~\cite{8851280}. 

Techniques can also be classified according to their ability to support out-of-sample or only in-sample scenarios. In-sample techniques use the entire dataset $\mathcal{X}$ to create the dimensionality reduction function $f$. Out-of-sample use a sample ${\mathcal{X'} \subset \mathcal{X}}$ to create ${f':\mathcal{X'} \rightarrow \mathcal{Y'}}$ and use $f'$ to map the entire (or the remaining of the) dataset. In principle, any linear technique can be used in an out-of-sample scenario since a linear transformation can be represented by a matrix. In this case, it is common to have approximations of in-sample techniques through a common approach of embedding a sample $\mathcal{X'}$ through an in-sample strategy followed by the interpolation of the remaining ${\mathcal{X}-\mathcal{X'}}$ data items. Examples involve Landmarks MDS~\cite{de2004sparse}, Landmarks ISOMAP~\cite{NIPS2002_5d6646aa}, and Sammon's approximation~\cite{DERIDDER19971307}. Some out-of-sample approaches are even agnostic and could be used to approximate any in-sample technique, such as the neural network-based strategies~\cite{doi:10.1177/1473871620909485, Dexter2020.09.03.269555}.

Finally, one last categorization, still not discussed in the literature, but useful to introduce in this paper, could involve the use or not of GD strategies. In fact, the idea that DR and GD are somewhat related has been pinpointed in the literature; for instance, the way t-SNE is discussed in the original paper uses the GD spring layout terminology, and UMAP explicitly declares that a nearest neighbor graph is constructed and embedded into the visual space when explaining the computational view of the technique. Indeed, the intersection between DR and GD is quite frequently considered by topology-based techniques, such as UMAP and TopoMap~\cite{9222271}. Our focus is to go beyond the introduction of a new categorization to split techniques, but to formalize and introduce a framework to support the identification and exploration of the opportunities yet to be explored.


\section{The Framework}

In the Introduction (see Section~\ref{se:intro}), we pose a high-level definition of DR as \textit{the process of mapping data items from a high-dimensional space into points in a lower dimensional visual space preserving the relationships between data items as much as possible on the mapped points}. From this definition, two key terms become central to our framework: \textbf{\textit{relationships}} (topology extraction) and \textbf{\textit{mapping}} (embedding generation). Relationships model \textit{what} the final layout intends to represent. The two most common are similarity and dissimilarity relationships. Mapping involves \textit{how} such modeling of relationships is transformed into visual representations through a spatialization strategy. 

Based on this decoupling, associating DR and GD domains is straightforward. Let ${\mathcal{G_D} = (\mathcal{V}, \mathcal{E}, \Omega)}$ be a graph representing the data set $\mathcal{X}$, where ${\mathcal{V}=\{v_1, \ldots, v_N\}}$ is a set of $N$ vertices, $\mathcal{E}$ is a set of unordered pairs $e_{ij} = (v_i, v_j, \omega_{ij})$ of vertices, called edges, and $\omega_{ij}: \mathcal{E} \rightarrow \mathbb{R}$ is a scalar associated with an edge $e_{ij}$. In our framework, each data instance $x_i$ corresponds to a vertex $v_i$, and the relationship between two data instances $x_i$ and $x_j$ is represented by an edge $e_{ij}$ between the vertices $v_i$ and $v_j$, in which the magnitude of such a relationship is defined by $\omega_{ij}$. Then, the DR process can be viewed as the embedding of such a graph into the visual space.

Our framework also includes \textit{\textbf{Quality Analysis}} as a way to validate the results of the relationships and mapping stages. 
Although it is not an integral part of DR techniques, quality analysis is one of the most important elements for DR analytical pipelines. Since DR techniques are approximations, it is usually not possible to capture all the datasets' traits through the modeled relationships and their mapping to the visual space. Hence, understanding the impact and extent of the information loss incurred in the process is essential to ensure whether the visual mapping is a reliable representation of the relationships and that the relationships reflect the dataset structure. The last stage of our framework, \textit{\textbf{Visualization \& Interaction}}, is another important element of any visualization pipeline where GD strategies can benefit DR. On a conceptual level, interaction can be defined as the interplay between humans and visualization interfaces involving a data-related intent~\cite{8805424}, so by considering our framework, existing graph-enhanced visual representations and advanced manipulations are candidates to support better representations and exploration of DR layouts.

Figure~\ref{fig:framework} illustrates our proposed framework. The backbone and theoretical basis of our framework show the different unified stages involved in going from data $\mathcal{X}$ to interactive visualization~\cite{mazza2009introduction}. \textit{Relationships} are constructed by defining a distance matrix (DR) or given from a defined topology (GD). There is a direct matching between the distances (similarity, dissimilarity) between data items and weighted graph connections. Note that a filled distance matrix matches with a fully connected graph. Using various techniques from GD, the standard distance matrix can be transformed to go beyond probabilities (see Section~\ref{sec:relationships}). After defining the relationships, a mapping is generated. Here there is a direct matching between embedding generation by defining a mapping from data $\mathcal{X}$ to points $Y$ in DR, and computing a layout in GD. This process, in fact, similarly defines a mapping from vertices $\mathcal{X}$ to points $Y$ in a lower dimensional space (typically $\mathbb{R}^{n=\{2,3\}}$). There are opportunities to apply mapping techniques from GD to DR (see Section~\ref{sec:mapping}). For both fields, the resulted layout is then compared against the (ground-truth) relationships to validate its quality. Also in this case, it is possible to outline a direct matching between quality metrics typically used in GD and DR, posing opportunities for hybrid utilization (see Section~\ref{sec:qualityanalysis}). Finally, \textit{Visualization \& Interaction} techniques enable a sense-making analytics loop. Also, here a matching between techniques could potentially benefit both fields but is considered out of scope as techniques are plenty and we aim to demonstrate that the fundamental stages before the visualization are similar, hence visualization and interaction techniques from both can be used interchangeably.   

When considering the two perspectives of DR and GD the different elements, techniques, and concepts involved for each can be matched to the associated stages. This makes explicit the similarities of both approaches and reveals ample opportunities for cross-fertilization between the two fields. In this paper, we focus on enhancing the DR field with concepts from the GD field. A similar exercise can be performed starting from the GD perspective, but this is left for future work (see also Section~\ref{sec:discussion}).

Next, we discuss how each of these stages is considered in the DR domain, give examples of how existing DR techniques can be viewed considering our association between DR and GD theory, and some opportunities for GD to help advance DR.



\subsection{Relationships}
\label{sec:relationships}
In light of the taxonomies discussed in Section~\ref{sec:related}, in our framework, local and global DR techniques can be viewed as having different goals with respect to the ``number'' of relationships that are aimed at being preserved. Global techniques can be modeled as weighted complete or fully connected undirected graphs (so-called pairwise relationships), while local techniques are sparser or modified representations (so-called neighborhood relationships). This can also be discussed in terms of the average degree of the nodes in $\mathcal{G_D}$. For global techniques, the average degree is equal to the number of nodes $N$ in $\mathcal{G_D}$, while for local techniques, this degree tends to be much smaller. 

Since most global techniques focus on preserving distances, in our framework, the weight of an edge $e_{ij}$ in $\mathcal{G_D}$ is set to be proportional to the distance $\delta_{ij}$ between the two data items $x_i$ and $x_j$ representing the vertices $v_i$ and $v_j$. Hence, for distance-based global techniques, in our framework $\mathcal{E} = \{(v_i, v_j, \delta_{ij})~|~(v_i, v_j, \omega_{ij}) \in V^2~\text{and}~v_i \neq v_j \}$. That is the case for MDS, IDMAP, and Sammon's Mapping techniques. PCA is another global technique that can also be conceptually represented in this way. Although the main focus of PCA is to create lower-dimensional representations preserving data variance, the resulting layouts are identical to those produced by MDS when $\delta$ is the Euclidean distance~\cite{van2009dimensionality}. So, at a conceptual level, PCA could also be modeled focusing on representing distance relationships.

Local techniques trim relationships or change edge weights, focusing on (small) neighborhoods. For example, LLE, LSP, and LoCH techniques use nearest neighbors to reduce relationships, only considering the $k$ more similar items to a data item to derive the relationships. In GD terms, the resulting graph only contains edges between nearest neighbors, rendering the well-known Nearest-Neighbor Graph (NNG). For local techniques based on nearest neighbors, in our framework ${\mathcal{E} = \{ (v_i, v_j, \delta_{ij})~|~v_i~\in~V, v_j~\in~\mathcal{N}_{v_i}\}}$, where $\mathcal{N}_{v_i}$ is the set that contains the nearest neighbors of $x_i$ (or $v_i$). The typical strategy used by, for example, LSP and LoCH to construct such NNGs is to connect every item $x_i$ to its $k$ most similar items. 


Considering neighborhoods and local relationships, some recent techniques have applied more sophisticated strategies to prune relationships. An example is TopoMap~\cite{9222271}. TopoMap uses minimum spanning trees to remove edges, focusing on preserving 0-homology, that is, the connection between groups of data items. UMAP~\cite{mcinnes2018umap} is another example. UMAP introduced the concept of ``fuzzy simplicial complexes'' to capture the topology of a dataset where an edge $e_{ij}$ and its weight $\omega_{ij}$ represent the likelihood (or probability) to exist, that is, the likelihood that items $x_i$ and $x_j$ are ``connected'' in a high-dimensional manifold. Notice that for UMAP, the weights $\omega_{ij}$ are not proportional to the distance between data items, but to the probability that they are neighbors in the high-dimensional space, so edge weights, in this case, represent how similar the vertices are instead of how dissimilar.

In addition to topological transformations (edge removal), some local techniques adopt a different strategy and change edge weights. An example is ISOMAP. The general idea of ISOMAP is to use geodesic distances to unfold manifolds. In this process, first, a k-NNG is created (this is an auxiliary structure and not our $\mathcal{G_D}$), and the geodesic distance $\delta'_{ij}$ is approximated as the length of the shortest path between $x_i$ and $x_j$ in this graph. Then, the full pairwise geodesic distance relationships are used to produce the final layout using MDS. In our framework, ISOMAP relationships are modeled as $\mathcal{E} = \{(v_i, v_j, \delta'_{ij})~|~(v_i,v_j, \omega_{ij}) \in V^2~\text{and}~v_i \neq v_j \}$. 

This conflicts with the typical definition found in the literature that local techniques are based on neighborhood relationships, in our framework modeled by graphs with lower average edge degrees, since, for ISOMAP, the average degree is equal to the number of vertices $N$. This requires a redefinition of local techniques to also consider the importance of the relationship instead of only the ``number'' of relationships. Incorporating, for local techniques, that relationships between far-apart data items are weighted to be much less important than relationships between neighbor data items. This differentiation between removing edges or changing their weights may seem minor, but it opens up many interesting opportunities later discussed.

Another well-known example of a local technique that could also be modeled as an edge-weight transformation strategy is t-SNE. t-SNE, as UMAP, focuses on representing neighborhood probabilities. It transforms distances into probabilities by applying a Gaussian function centered on each data item $x_i$ on the high-dimensional distances $\delta_{ij}$. To model t-SNE in our framework, let $p_{j|i} = \nicefrac{exp(-\delta_{ij}/2\sigma_i^2)}{\sum_{k\neq i}exp(-\delta_{ik}/2\sigma_i^2)}$ be the conditional probability of $x_j$ being a neighbor of $x_i$, with $p_{ij}=\nicefrac{(p_{j|i}+p_{i|j})}{2}$ the symmetrized conditional probability, then $\mathcal{E} = \{(v_i, v_j, p_{ij})~|~(v_i,v_j, \omega_{ij}) \in V^2~\text{and}~v_i \neq v_j \}$. In t-SNE, the variance $\sigma_i$ is controlled using a parameter called perplexity, which can be interpreted as a smooth measure of the effective number of neighbors used in the process~\cite{van2008visualizing}. Notice that t-SNE could also be modeled as an edge removal strategy if edges with probability near zero are removed, that is, $\mathcal{E} = \{ (v_i, v_j, p_{ij})~|~v_i~\in~V, v_j~\in~\mathcal{N}_{v_i}\}$, where $\mathcal{N}_{v_i} = \{ v_j~|~v_i~\neq~ v_j \text{ and}~p_{ij}~\gtrapprox~0\}$. 
For illustration, Table~\ref{tab:tech_relation} lists some of the techniques discussed in Section~\ref{sec:related} and the associated modeled relationships.

\begin{table}
    \centering
    \caption{Examples of DR techniques and the modeled relationships.}
    \resizebox{\columnwidth}{!}{%
    \begin{tabular}{p{1.5cm}|p{8cm}}
        \textbf{Technique} & \textbf{Modeled Relationships} \\
        \hline
         t-SNE & Joint probability of data items to be neighbors.\\
         UMAP & Probability of data items to be neighbors.\\
         ISOMAP & Geodesic distance given a nearest neighbor graph.\\
         MDS & Pairwise high-dimensional distances.\\
         Sammon's Mapping & Pairwise high-dimensional distances.\\
         LAMP & Pairwise weighted high-dimensional distance between data items in a sample and high-dimensional distance to these samples and other items in the dataset.\\
         IDMAP &  Pairwise high-dimensional distance.\\
         LSP &  Weighted high-dimensional distance between nearest neighbors.\\
         ProjClus  &  Pairwise high-dimensional distances between data items in the same cluster\\
         LLE  &  Weighted high-dimensional distance between nearest neighbors.\\
         LoCH  &  Weighted high-dimensional distance between nearest neighbors.\\
         Laplacian Eigenmaps & Pairwise distance between nearest neighbors.\\
         PLSP & Weighted high-dimensional distance between nearest neighbors\\
    \end{tabular}
    }
    \label{tab:tech_relation}
\end{table}

Some of the most recent local DR techniques focus heavily on using graph theory to capture data topology and represent local structures. However, there are still opportunities in graph theory that remain unexplored and could benefit DR processes. Next, we discuss some ideas.

\noindent\textbf{Opportunities} GD theory goes beyond fully connected and nearest-neighbor graphs and presents some interesting ideas that can be used to model the relationships. One direction to potentially better preserve the global structure in DR, is to consider the weight matrix as relationships and compute a weight-aware backbone from this using GD techniques. The backbone could, for example, be the minimum-spanning tree $T$~\cite{Prim1957, Kruskal1956}, where $T$ is a tree, connects all vertices/data items, and, the sum of edge weights $\omega_{ij}$ is minimized. The distances corresponding to the relationships involved in the resulting spanning tree $T$ should then be strengthened, while others are decreased. To also preserve local structure, there is an opportunity to explore the use of graph spanners as a backbone, preserving the lengths of the shortest paths in $G_\mathcal{D}$ up to some amount of distortion~\cite{Ahmed2020}. It would be interesting to explore the relation between the quality of the resulting DR and the amount of allowed distortion.

Other opportunities for scaling selected distances to strengthen relationships are to determine what relationships should be scaled based on graph-theoretic metrics, such as centrality~\cite{Newman2010}. Centrality measures compute node ranking according to the importance of interest. Typically, the metrics are an indication of key nodes in the network. Two notable measures to explore are \emph{closeness}~\cite{Bavelas1950}, and \emph{betweenness}~\cite{Freeman1977}. The closeness of the node indicates the average length of the shortest path from the node to all other nodes in the network; therefore, the higher the score, the closer this node is to all other nodes. The betweenness score is based on the number of shortest paths that cross the node. If high-betweenness nodes are removed from a graph, this potentially results in disconnected components. This knowledge can potentially be used to either strengthen the relationships involved for a better global layout, or diminish for a better local layout (or finding a balance between the two). Also, since centrality measures identify the key nodes in a graph, the relationships involved between any two key nodes can be used as a backbone. Graph-based distances could also be used to transform global relationships into local ones by changing edge lengths, such as the Shared Nearest Neighbors (SNN) distance~\cite{Jarvis1973ClusteringUA}. Potentially, distance transformation and the combination of multiple measures in a multi-criteria optimization scheme seem advantageous; however, which ones and how to combine these is an open question.

Using graph-theoretic measures for a multi-step DR approach poses another interesting area for exploration. First, finding (data facet) communities using community detection (e.g., using the Louvain~\cite{Blondel2008} or Leiden~\cite{Traag2019} methods), and laying out the components, the components can then be repositioned to reflect global distances. Removing vertices is also possible and would result in not a topological transformation but an aggregation. In DR, some techniques use such a concept for multi-level exploration, navigating between more and less detailed representations. One example is HiPP~\cite{Paulovich2008}, a hierarchical point-placement approach for the exploration of multi-dimensional datasets. In general, graph reduction techniques~\cite{ijcai2024p0891} could enable multi-step DR approaches, laying out a reduced set of vertices first. Three techniques to reduce graphs are \emph{sparsification} identifying significant nodes, \emph{coarsening} by aggregating similar nodes and edges (e.g., graph summarization~\cite{10.1145/3186727}), and \emph{condensation} that learns a synthetic graph from scratch.

%

\subsection{Mapping}
\label{sec:mapping}
Following our framework, once the relationship graph ${\mathcal{G_D} = (\mathcal{E}, \mathcal{V}, \Omega)}$ is modeled, it is embedded in the visual space. DR techniques generate an embedding using different strategies and approaches, usually using either analytical or numerical solutions. In this section, we refrain from discussing analytical solutions, such as the process applied by the classical MDS algorithm, and will focus on numerical strategies since there are more opportunities to benefit DR from the extensive GD theory.

Typically, numerical solutions are based on optimizing a cost function, which is usually performed using some gradient-descent-like strategy. At a high level, the relative position of a point $x_i$ in a DR layout $\mathcal{Y}$ is defined based on the position of other points in $\mathcal{Y}$ and the relationships $x_i$ has with the other items $x_j$. Using our framework terminology, this is achieved by positioning two vertices $v_i$ and $v_j$ on the visual space so that the distance (usually Euclidean) between them is proportional (or inversely proportional) to the weight $\omega_{ij}$ of the edge $e_{ij}$ between them. Using the well-known forces metaphor, during the mapping process, given an edge $e_{ij}$, a vertex $v_i$ is attracted to $v_j$ if the distance between them is larger than $\omega_{ij}$, otherwise $v_i$ is repelled by $v_j$. If $v_i$ is not connected to $v_j$, $v_j$ will repulse $v_i$ if it gets closer than expected. That is, attracted if they are far apart or repulsed if they are closer than they should be. 

A classical example of global techniques employing such a strategy is Sammon's Mapping (SM) technique~\cite{sammon1969nonlinear}. SM uses a quasi-Newton method, requiring both the gradient and the second derivative of its cost function. It is beyond the scope of this paper to dissect the concepts of function optimization, but the basic concept of gradient descent-like strategies is to optimize a cost function by moving in the opposite direction of its gradient. Using our framework, the SM gradient can be expressed as follows:

\begin{equation}
\dfrac{\delta}{\delta v_i} = \dfrac{-2}{\sum_i \sum_{j>i} d_{ij}} \sum_i \sum_{j > i} \left(\dfrac{d_{ij}-\omega_{ij}}{d_{ij} \omega_{ij}} \right) (v_j - v_i)
\end{equation}

Intuitively, through this optimization $v_i$ moves toward $v_j$ when the distance between them ($d_{ij}$) is larger than the distance between the data elements they represent ($\omega_{ij} = \delta_{ij}$), and is repelled otherwise. IDMAP~\cite{minghim2006idmap}, another global technique, uses a similar principle, but instead of moving $v_i$ in the direction of all other vertices $v_j$ in one loop, moves all other vertices $v_j$ in the direction of $v_i$. In turn, this adds much more energy into the system, usually converging with fewer iterations.

Gradient descent-like approaches have also been extensively used for local techniques, but to work, they usually require additional steps. An example of a local mapping process that presents limitations is the one used by the original~\cite{NIPS2002_6150ccc6} and symmetric SNE technique~\cite{van2008visualizing}. Symmetric SNE relationships are the same as those of the t-SNE, representing the probability ($\omega_{ij} = p_{ij}$) that two data items $x_i$ and $x_j$ are neighbors in the high-dimensional space. In the mapping process, the idea is to calculate the joint probability $q_{ij}$ of the respective points (vertices) $v_i$ and $v_j$ to be neighbors in the visual space, and move these vertices so that $q_{ij}$ matches $p_{ij}$ in the sense of the Kullback-Leibler (KL) divergence. The gradient of this optimization is as follows:

\begin{equation}
    \dfrac{\delta}{\delta p_i} = 4 \sum_j (p_{ij}-q_{ij})(v_i-v_j)
\end{equation}

So, $v_i$ moves toward $v_j$ (in the opposite direction of the vector $\vec{v}=(v_i-v_j)$) when $p_{ij}-q_{ij} > 0$, that is, when the probability in the high-dimensional space is higher than in the visual space. The issue with this solution is that considering the probabilistic transformation employed by the symmetric SNE, the attraction forces dominate the optimization. Recall that symmetric SNE (as the original SNE) focuses on preserving relations between neighbors, so if $x_i$ is reasonably dissimilar (not among the nearest neighbors) from $x_j$, but $v_j$ is not between the nearest neighbors of $v_i$, $(p_{ij}-q_{ij}) \approx 0$, $v_i$ is not attracted by $v_j$ (as expected) but also not repelled (as it should). In other words, the amount of repulsion forces is reasonably small and (some) dissimilar data items tend to be crushed in the center of the visual space, resulting in the ``crowding problem''. t-SNE presents an elegant solution to this problem, replacing the Gaussian distribution by the t-student distribution to model probabilities in the visual space. The new probability and the resulting gradients are as follows:

\begin{equation}
    q_{ij} = \dfrac{(1 + ||v_i - v_j||^{2})^{-1}}{\sum_{k \neq l} (1 + ||v_k - v_l||^{2})^{-1}}
\end{equation}
\begin{equation}
    \dfrac{\delta}{\delta p_i} = 4 \sum_j (p_{ij}-q_{ij})(v_i-v_j)(1 + ||v_i - v_j||^{2})^{-1}
\end{equation}

The relationships between the data items do not change, but the neighborhoods in the visual space become larger, introducing more repelling forces between dissimilar data items that are close in the visual space compared to the symmetric (and original) SNE methods~\cite{van2008visualizing}.

The idea of adding extra repulsive forces to improve layouts based on neighborhoods has also been explored by other DR methods. UMAP is one example~\cite{mcinnes2018umap}. In the UMAP optimization process, instead of changing the optimization gradient, repulsive forces are added between $v_i$ and random sample points (vertices) that are not in its neighborhood, that is, not connected to $v_i$. This is called negative sampling, and it was shown to improve the produced layouts. Similarly, the Barnes-Hut version of t-SNE~\cite{JMLR:v15:vandermaaten14a} adds forces between $v_i$ and the centroid of groups of points not connected to $v_i$ to model repulsive forces and also to speed up the entire process. Other strategies have also been suggested to improve this process, such as the ``early exaggeration'' on t-SNE, which boosts attraction forces by multiplying the probabilities between data items by a constant in the initial optimization iterations. 


The idea of better modeling the attraction and repulsive forces of DR techniques based on neighborhood relationships has recently been considered by Bohm et al.~\cite{JMLR:v23:21-0055}, placing techniques in a spectrum of such forces. In the paper, the authors derive a new gradient formulation for t-SNE where the balance of attraction and repulsion forces can be explicitly controlled (similar to the early exaggeration), emulating other local DR techniques, including UMAP, by carefully controlling the trade-off between attraction and repulsion forces. Good evidence is presented supporting the fact that the real difference between t-SNE and UMAP is the mapping phase, since the relationships are very similar and the gradient of both becomes identical under parameterization (we later discuss this with examples in Section~\ref{se:experiments}). The implication is that the typical condensed clusters found in UMAP layouts and their argued superior quality in representing global structures are the result of negative sampling (graph drawing), allowing one to conclude that the stated and effective loss functions are qualitatively different~\cite{JMLR:v23:21-0055}.

In summary, while for global techniques attraction and repulsion forces are part of the optimization since relationships are modeled by a fully connected graph, on local, repulsion forces need to be boosted to improve separation between natural clusters to form. Therefore, by changing the strategy used to model such forces in local techniques, perceptual visual results are drastically affected.




\noindent\textbf{Opportunities.} Modeling the idea of attraction and repulsion forces has long been discussed by the GD community, and is usually called \textit{force-directed placement} (please refer to the survey by Cheong et al. for an extensive treatment~\cite{cheong2020force}). Force-directed (FD, or energy-based) algorithms have two constituting elements: an energy model, encapsulating how nodes interact with each other, and an energy minimization strategy. The most basic energy model works by treating the nodes and edges of the graph as elements of a physical system. Vertices become same-sign charged particles (that tend to repel) and edges become springs, pulling the vertices at their ends close~\cite{eades84, fruchterman1991graph}. Other methods only use spring forces, whose ideal length matches the graph-theoretic distance between nodes~\cite{kamada1989algorithm}. The energy minimization strategy iteratively reduces the residual energy within the system until a (local) minima is reached. This is matched by a condition of balance between the forces, leading to the final configuration (or layout) of the nodes in the plane. 

This family of techniques produces good quality results but for small graphs only (hundreds of nodes), due to the poor scalability and tendency to get stuck into local minima of the energy function. \textit{Multi-level} FD was introduced to overcome this limitation. The layout process was divided into two phases: coarsening, where a hierarchy of simpler versions of the original graph is generated, and refinement. At this stage, and starting from the coarsest one, each graph in the hierarchy is drawn using a FD algorithm. The resulting coarse layout is used to place the vertices at the level below, leading to faster and better quality convergence due to a more favourable initial condition than random placement. More and more sophisticated multi-level FD algorithms have been introduced over the years~\cite{cheong2020force}.
Refinement~\cite{gayer2004grip} is achieved by ``transferring" the gradient values of nodes from the coarser levels to the finer ones, and using this favorable starting condition to achieve quicker and better convergence. This is an exquisite example of how a graph drawing optimization technique could boost the scalability of DR techniques following our suggestion of modeling relationships as graphs. 

Thanks to the modularity of the multi-level acceleration technique, there are several further opportunities to leverage this methodology. One example, drawing from DRGraph, is to improve and experiment with different coarsening techniques. Bartel et al.~\cite{bartel2010multi} conducted an experimental evaluation where they compared the performance of different coarsening strategies in terms of the number of levels produced. A good coarsening strategy should not generate more levels than necessary (more levels slow down the drawing process). The Solar Merger by Hachul and J{\"{u}}nger~\cite{hachul2004fmmm} produced the least amount of levels across the entire set of the benchmark graphs. It works by first ordering nodes by degree and, in descending order, it merges together each node with its neighbors up to distance 2. This information could be used in the refinement stage, to further optimize the refinement algorithm.

It is interesting that the problem that t-SNE solves compared to the symmetric (original) SNE method, if translated to GD based on our framework, finds some parallels to strategies suggested for GD. t-SNE addressed the so-called ``crowding problem'' by boosting the repulsion forces by calculating the attraction forces between neighboring (connected) nodes and the repulsion between all other nodes~\cite{eades84}. An interesting opportunity in this context would be to adopt an energy model whose minimum energy layouts naturally represent the cluster structure. Noack~\cite{noack2007energy} presented two energy models in this sense: node- and edge-repulsion \textit{LinLog}. These models improve cluster separation and could be a potential opportunity to deal with the crowding problem.
As we discussed in the previous paragraph, however, this strategy is slow due to the number of pairwise forces to compute. In GD, the use of the Barnes-Hut approximation~\cite{10.1007/3-540-44541-2_19} (i.e., space decomposition to simplify forces calculation) was already known at least one decade before it was suggested as a strategy to speed up t-SNE~\cite{JMLR:v15:vandermaaten14a}. Such strategy is currently used in the most advanced GD techniques (e.g.,~\cite{hu2005efficient, jacomy2014forceatlas2}). Similar happens with the strategy employed by UMAP, called negative sampling. In GD, it is called random vertex sampling~\cite{10.1111/cgf.13724}, and is known to produce similar results compared to Barners-Hut in a fraction of time.

\subsection{Quality Analysis}
\label{sec:qualityanalysis}
Dimensionality Reduction (DR) is, per definition, a lossy process. Unless the intrinsic dimensionality of ${\mathcal{X} \in \mathbb{R}^m}$ is equal to or lower than the visual $\mathbb{R}^p$, information will be lost in the process. In other words, some of the modeled relationships will not be represented or fully represented in the produced visual layouts. Therefore, an essential element of any analysis based on DR is to validate the faithfulness of the produced layouts, that is, how much the visual representation captures the intended data relationships.

Usually, quality analysis or validation involves metrics to evaluate the visual representation quality or interpret the preserved relationships on the produced layout. These measure how much of the similarities in the high-dimensional space are preserved and help interpret the preserved relationships on the produced layout. Stress and neighborhood preservation, and all their derivations, are typically the most common metrics for DR layouts validation. The stress measures the difference between the relative distance between two points in the low-dimensional representation and the same points in the original high-dimensional space~\cite{kruskal1964multi}. Other distance error metrics include correlation~\cite{borg2005modern} and silhouette coefficients~\cite{mining2006introduction}. Neighborhood preservation, on the other hand, evaluates whether the \textit{k}-nearest neighbors of each data point correspond in the high-dimensional space and low-dimensional representation~\cite{venna2001neighborhood, FADEL2015546}. 
Trustworthiness is a reliable metric as it both indicates whether the neighborhood composition has changed during the projection, but also quantifies whether the neighbor \textit{order} relative to each point did~\cite{venna2001neighborhood}. If ground truth (i.e.,  node labeling) is available, Neighbor Hit~\cite{4378370} captures how much the grouping defined in the labels is preserved in the visual layout.

Although it is sometimes common to go blind and use all or most of the metrics to validate a layout, considering our DR framework segmentation into \textit{Relationships} and \textit{Mapping}, it is possible to shed some light on what these metrics are, in fact, measuring. Is it the quality of the modeled relationships? Is it the quality of the spatialization or embedding into the visual space? Or a combination of both? Neighborhood preservation techniques focus on the former and provide an indication of how the mutual relationships between the data points have been preserved in the low-dimensional representation. On the other hand, distance error measures evaluate whether the mapping stage was accurate; that is, if the relative positioning of the nodes in the low-dimensional projection reflects the distances in the original high-dimensional space.

An interesting aspect of splitting the process into relationships and mapping is the ability to understand and have a better lead when setting the parameters of a DR technique. This can be achieved by splitting the parameters into relationship and mapping parameters and using proper metrics to measure the quality of these elements. Consider t-SNE as an example; the distance metric and perplexity can be viewed as relationship parameters since they change how the probabilities are computed. Early exaggeration, learning rate, and number of iterations are mapping parameters since they basically influence the gradient descent process and do not change the modeled relationships. In other words, it is possible to understand better what is at fault if the layout does not reflect what is expected. Furthermore, this allows, for instance, to assess the real differences between techniques, for instance, the differences between t-SNE and UMAP (sometimes claimed to be very similar~\cite{JMLR:v23:21-0055} or different~\cite{mcinnes2018umap}), helping to answer what is similar/different, relationships or the mapping?

\noindent\textbf{Opportunities.} In GD, there is an abundance of metrics that characterize both the topology of a graph (e.g., centrality measures, clustering coefficients, etc.) and the characteristics of a drawing (e.g., edge length, number of crossings, etc.). Centrality measures allow the user to rank nodes and edges depending on their position within the network, representing whether they are highly connected to the rest of the nodes or are placed within a tight cluster with few external connections. This information could also be used to validate the mapping, providing a deeper understanding of the results produced by t-SNE or UMAP. We provide an example of how this could be achieved in Section~\ref{se:experiments}.

\textit{Shape-}based metrics, introduced by Eades et al.~\cite{eades2017similarity}, provide a methodology to validate whether a graph layout geometry, obtained through a projection or a layout algorithm, is \textit{faithful} to the original topology of the graph. The premise is simple: a faithful drawing places close together two nodes in the plane if the same nodes are close in the original topology. A way to measure whether this property is satisfied or not is as follows. A shape (proximity) graph is generated from the coordinates of the nodes in the plane. Examples of shape graphs are the k-NN graph, Delaunay triangulations, or the Gabriel Graph. The shape graph is then compared to the original topology by evaluating the Jaccard similarity of the two edge sets. The same process could be applied to a low-dimensional projection obtained through DR. If a high similarity is achieved, then the resulting projection is faithful to the original topology. 

Another opportunity lies in understanding the differences between two different projections, say between t-SNE and UMAP. This could be achieved with a form of edge matching: Jung et al.~\cite{10360924} in their paper investigate the problem of identifying robust and re-occurring subgraphs in multidimensional projections. Once the corresponding k-NN graphs on the projections are generated, subgraph pattern matching~\cite{jiang2013survey} is applied to identify recurring structures. Network motifs could be used for the same purpose. A motif is a recurring pattern found (potentially multiple times) in complex networks: typically, they have the topology of stars, fans, or cliques~\cite{dunne2013motif}. The number and distribution of the motifs present in the network provide meaningful insight into the topology of the network. In turn, this can lead to a better understanding if the resulting projection still retains the structures that are present in the high-dimensional space - thus validating the mapping process.

\section{Experiments}\label{se:experiments}

In this section, some experiments are discussed to show the application of the proposed framework. It is not our goal to cover extensively the literature or quantitatively compare techniques. Our aim is to shed some light on the possibilities our framework can open. In these experiments we use the \textbf{digits} dataset~\cite{digits_data}, composed of bitmaps of handwritten digits from a pre-printed form. It contains $1,797$ data items in $64$ dimensions of $10$ different classes. 

\begin{figure}
\begin{subfigure}{.5\linewidth}
  \centering
  \includegraphics[width=\linewidth]{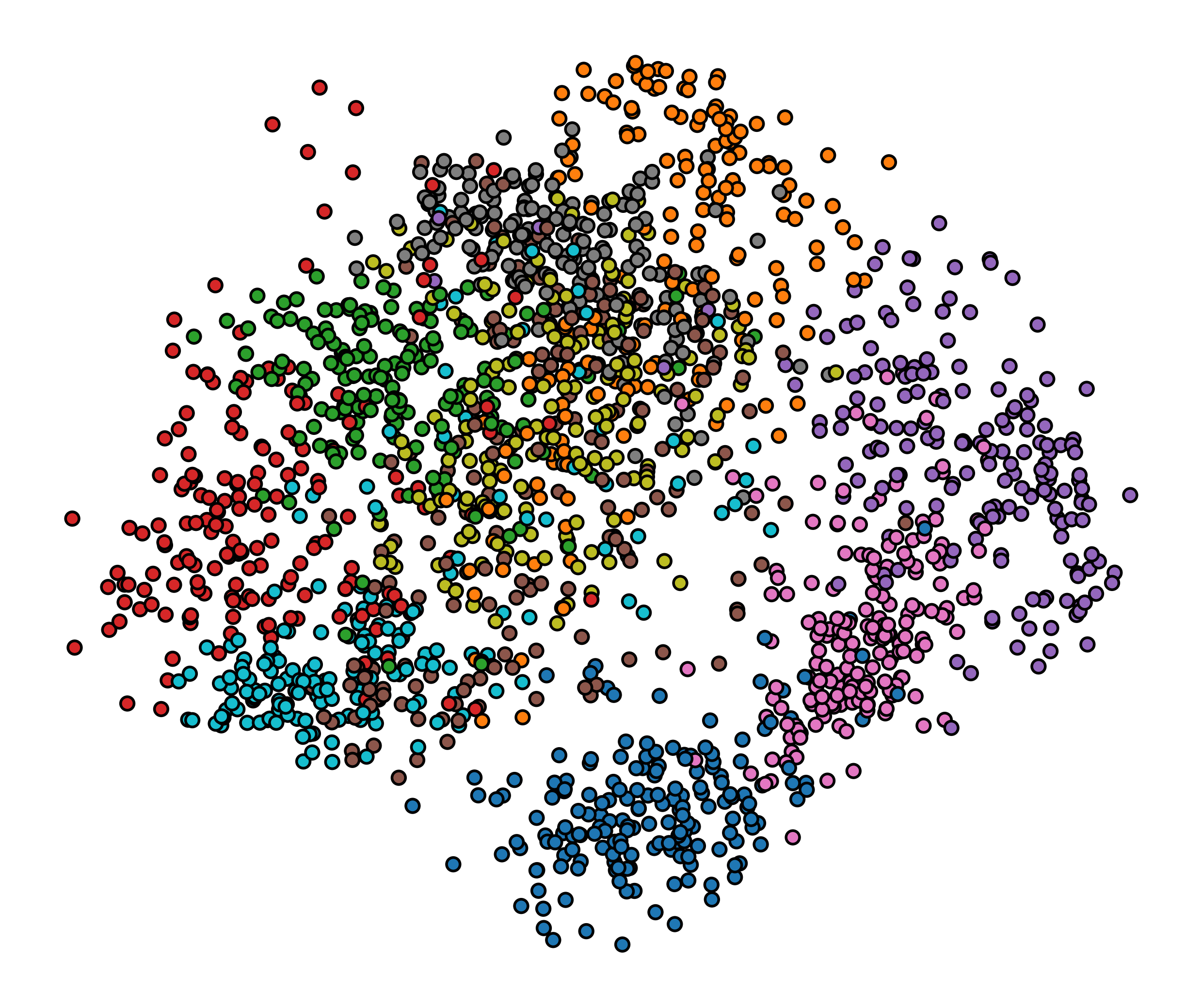}
  \caption{Classical MDS.}
  \label{fig:cmds}
\end{subfigure}%
\begin{subfigure}{.5\linewidth}
  \centering
  \includegraphics[width=\linewidth]{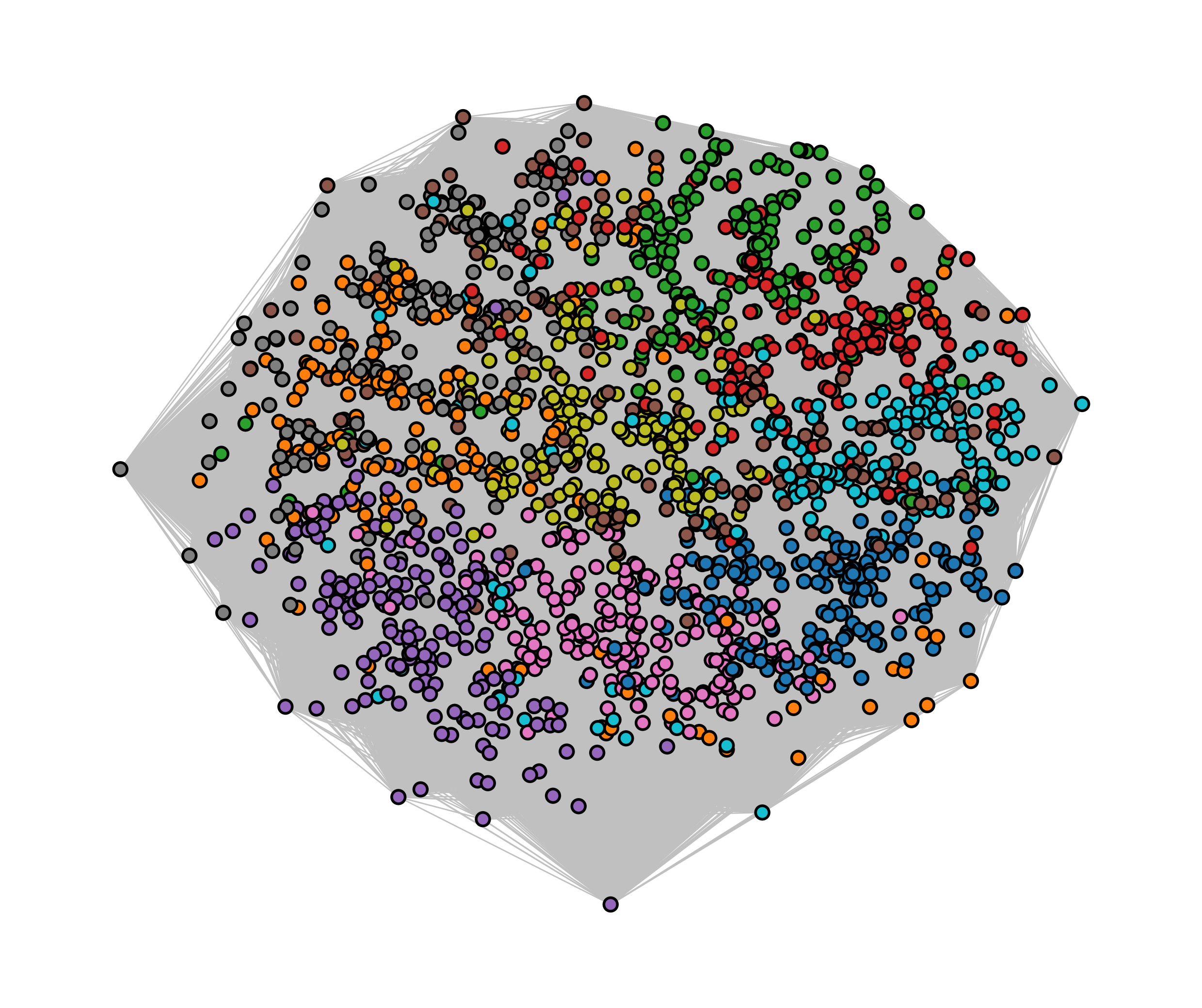}
  \caption{Fully connected graph.}
  \label{fig:pairwise}
\end{subfigure}
\caption{Comparing layouts created using the classical MDS technique and the spring layout of a fully connected distance graph. Both layouts fail to separate the existing classes, indicating modeling pairwise (global) relationships is not indicated if the goal is to see clusters.}
\label{fig:global}
\end{figure}

In the first experiment, we show different global layouts for this dataset. Figure~\ref{fig:cmds} shows the layout produced by the classical MDS technique, and Figure~\ref{fig:pairwise} shows the layout of a fully connected graph in which the edges are the (Euclidean) distances between the data items. This graph was drawn using the Fruchterman and Reingold algorithm, also known as the spring layout technique (see Section~\ref{sec:mapping}). Since such an algorithm works with similarity instead of dissimilarities, we change the edges' weights of this graph to $\omega_{ij}=(1 - \omega_{ij}/\omega_{max})$, where $\omega_{max}$ is the maximum edge weight in the original graph. Although some groups are somewhat visible in both layouts, e.g., the purple, blue, and pink groups, the global modeling of the relationships fails to separate the classes, which seems to not depend on the employed mapping approach but more on the modeled relationships since the MDS technique uses an analytical strategy that guarantees to best preserve the pairwise distances between data items under the Frobenius norm for laying out points. 

\begin{figure}[htb]
\begin{subfigure}{.5\linewidth}
  \centering
  \includegraphics[width=\linewidth]{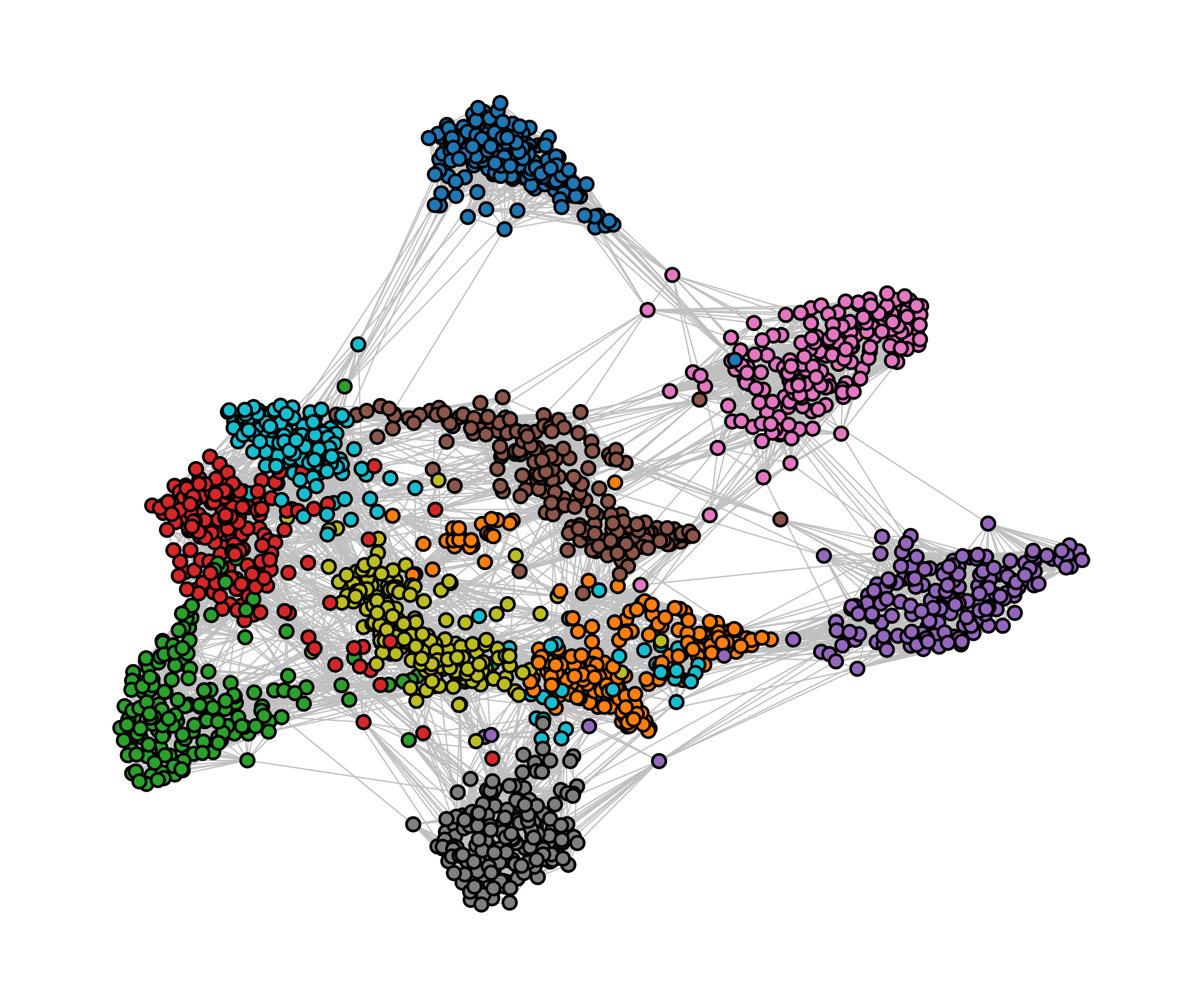}
  \caption{k-NN graph.}
  \label{fig:knn}
\end{subfigure}
\begin{subfigure}{.5\linewidth}
  \centering
  \includegraphics[width=\linewidth]{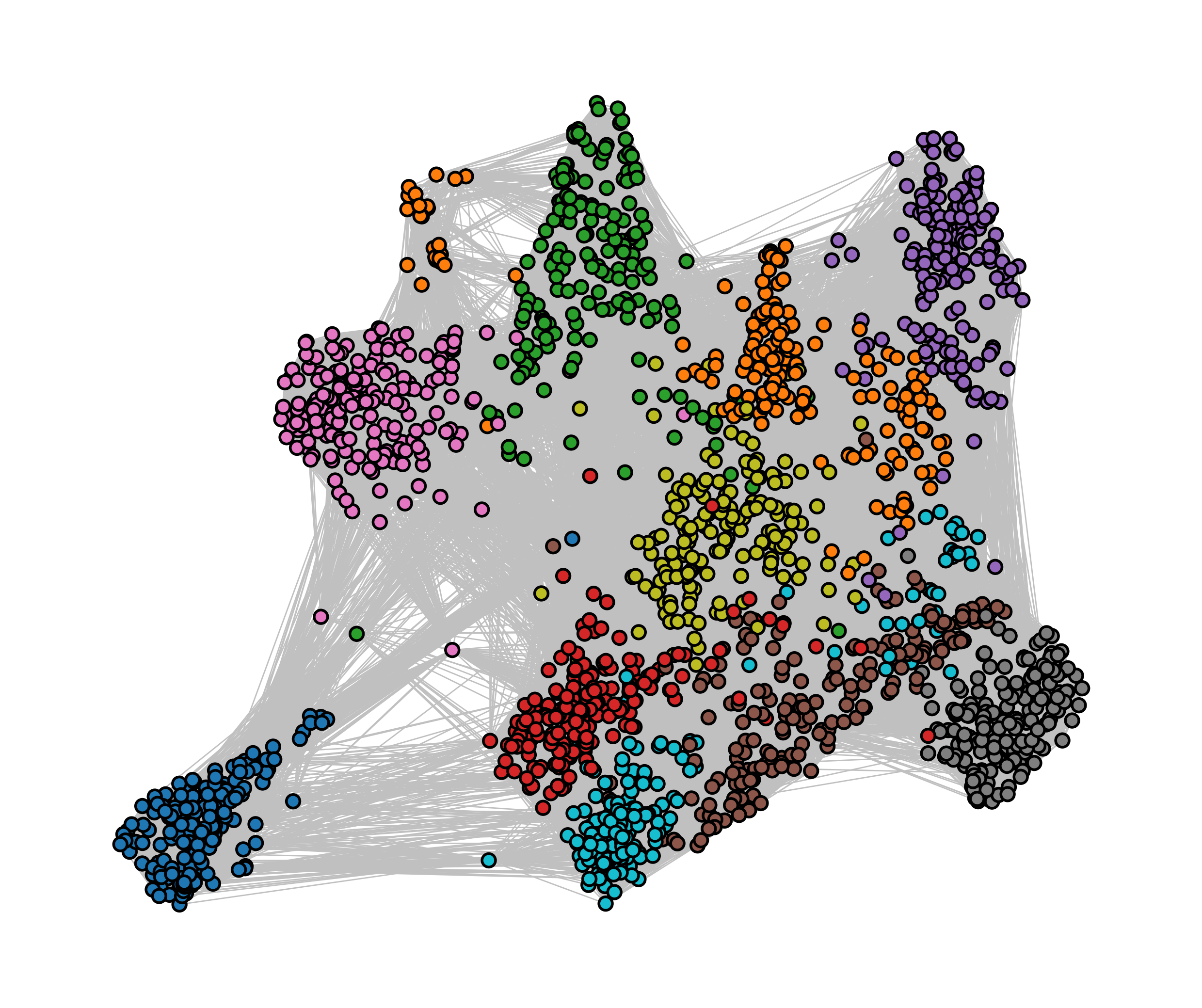}
  \caption{SNN graph.}
  \label{fig:snn}
\end{subfigure}%
\caption{Graphs modeling local relantionships. Compared to the previous example (Figure~\ref{fig:global}), local relationships are capable of better spliting the groups of data items with same label, suggesting that local reltionships are prefered for group segregation.}
\label{fig:global}
\end{figure}

The usual approach to improve such results is to apply some sort of manifold learning strategy, typically resulting in modeling local relationships. As discussed, local relationships in our framework are obtained by transforming global relationships using two different approaches. Either by removing or re-weighting the modeled edges (see Section~\ref{sec:relationships}). Figure~\ref{fig:knn} shows the result of layouting a k-Nearest Neighbor Graph (k-NNG) using the spring layout algorithm. In this example $k=10$, so every data item is connected to its $10$ most similar data item, and again $\omega_{ij}=(1 - \omega_{ij}/\omega_{max})$. We also applied an edge weight transformation approach to modify the modeled relationships. The result is shown in Figure~\ref{fig:snn}. In this example, to transform the edge weights, we use the Shared Nearest Neighbors (SNN) similarity~\cite{Jarvis1973ClusteringUA}. The SNN similarity $snn(x_i, x_j)$ between two data items $x_i$ and $x_j$ is defined as the number of nearest neighbors they have in common. So in the resulting graph, the edge weights are transformed to $\omega_{ij}=snn(x_i, x_j)$. Again, the graph is drawn using the spring layout technique using the same parametrization of the previous example to keep consistency -- throughout this section, parameters for the spring layout algorithm are always kept the same, and the initialization is performed using PCA. In both, the separation between groups improved considerably, also suggesting that the usually good results obtained in terms of group segregation are indeed the result of the modeled relationships since the mapping phase is the same as used in the previous example. 

\begin{figure}
\begin{subfigure}{.5\linewidth}
  \centering
  \includegraphics[width=\linewidth]{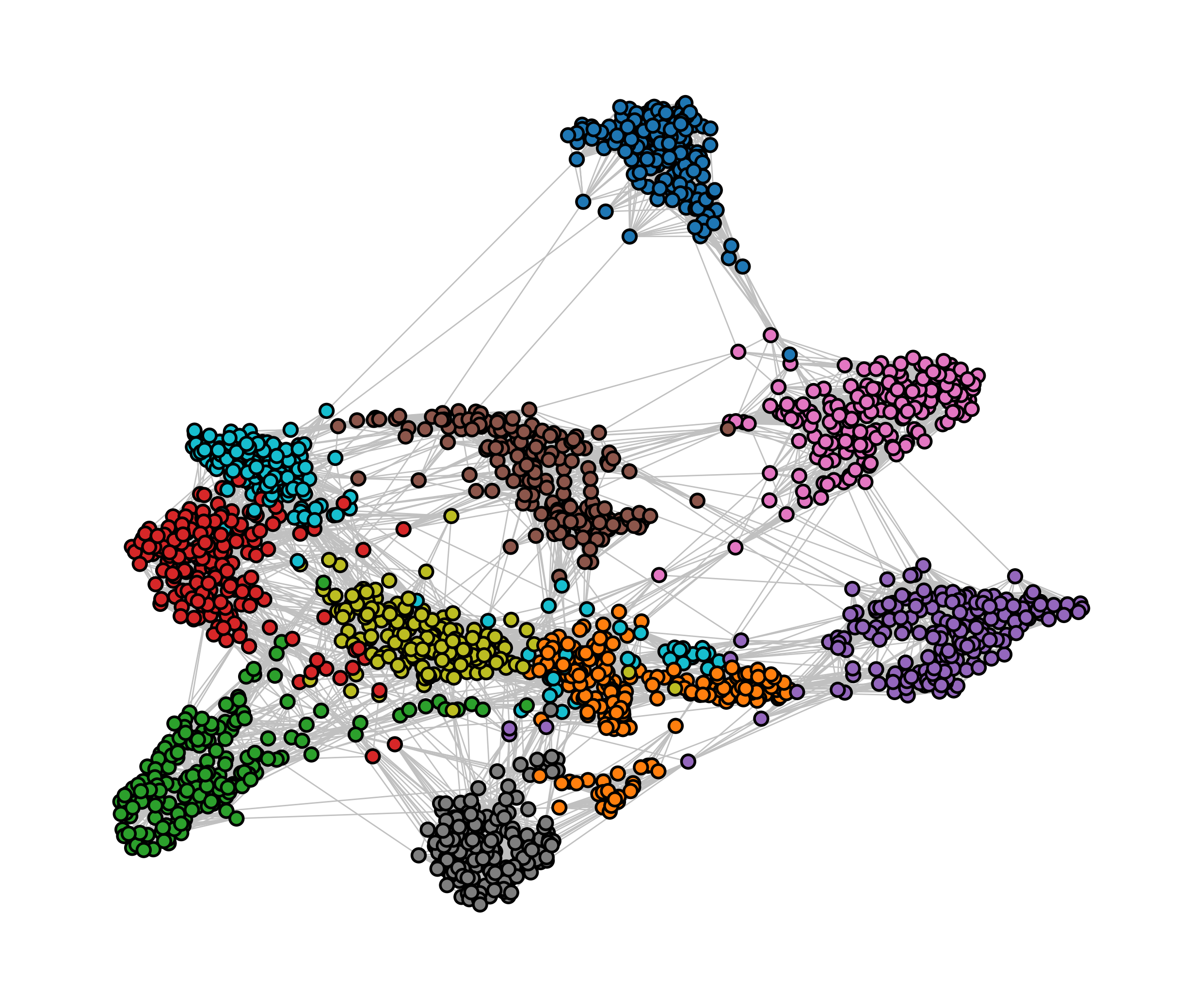}
  \caption{t-SNE graph.}
  \label{fig:gd_tsne}
\end{subfigure}
\begin{subfigure}{.5\linewidth}
  \centering
  \includegraphics[width=\linewidth]{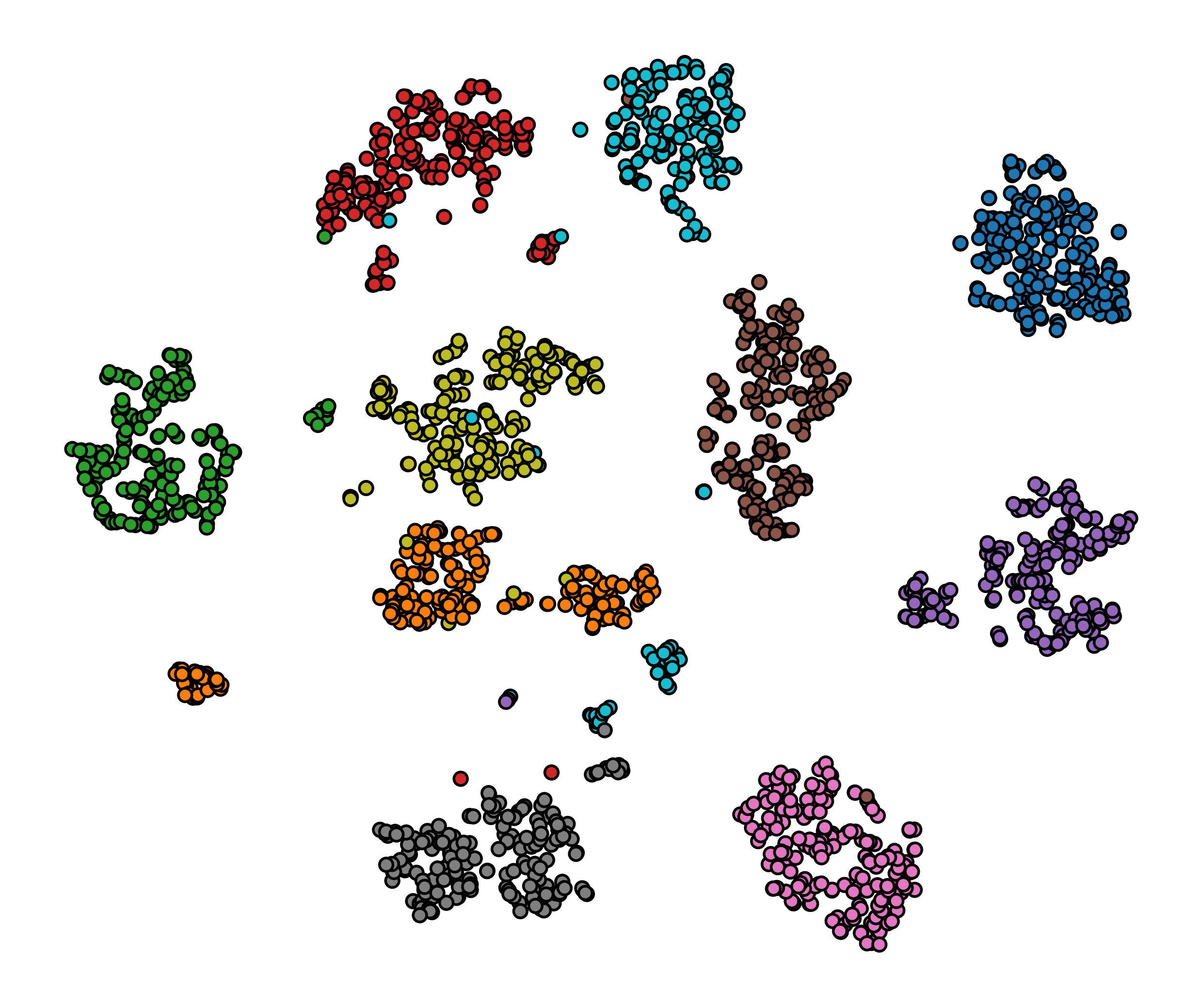}
  \caption{Original t-SNE.}
  \label{fig:tsne}
\end{subfigure}\\
\begin{subfigure}{.5\linewidth}
  \centering
  \includegraphics[width=\linewidth]{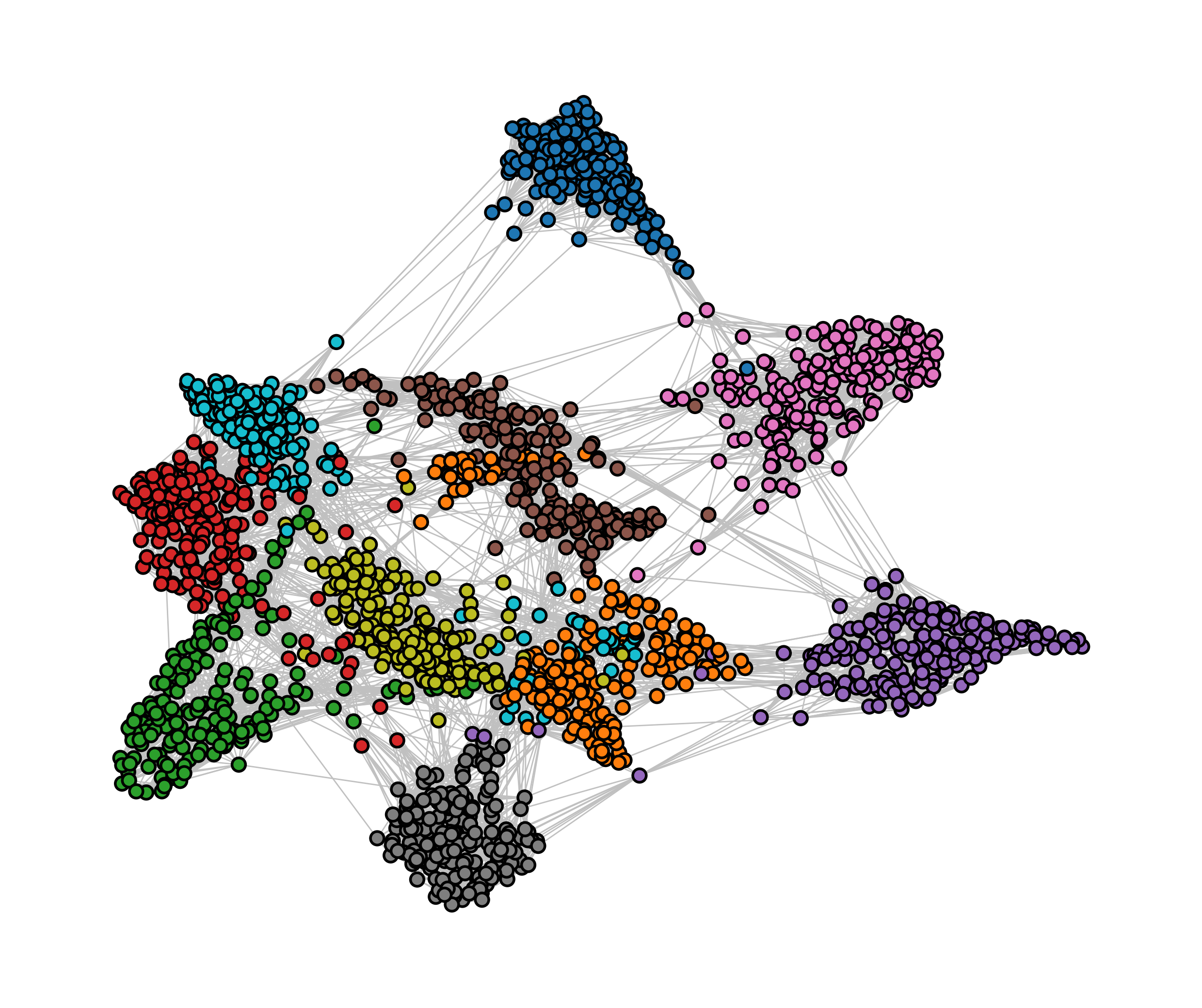}
  \caption{UMAP graph.}
  \label{fig:gd_umap}
\end{subfigure}
\begin{subfigure}{.5\linewidth}
  \centering
  \includegraphics[width=\linewidth]{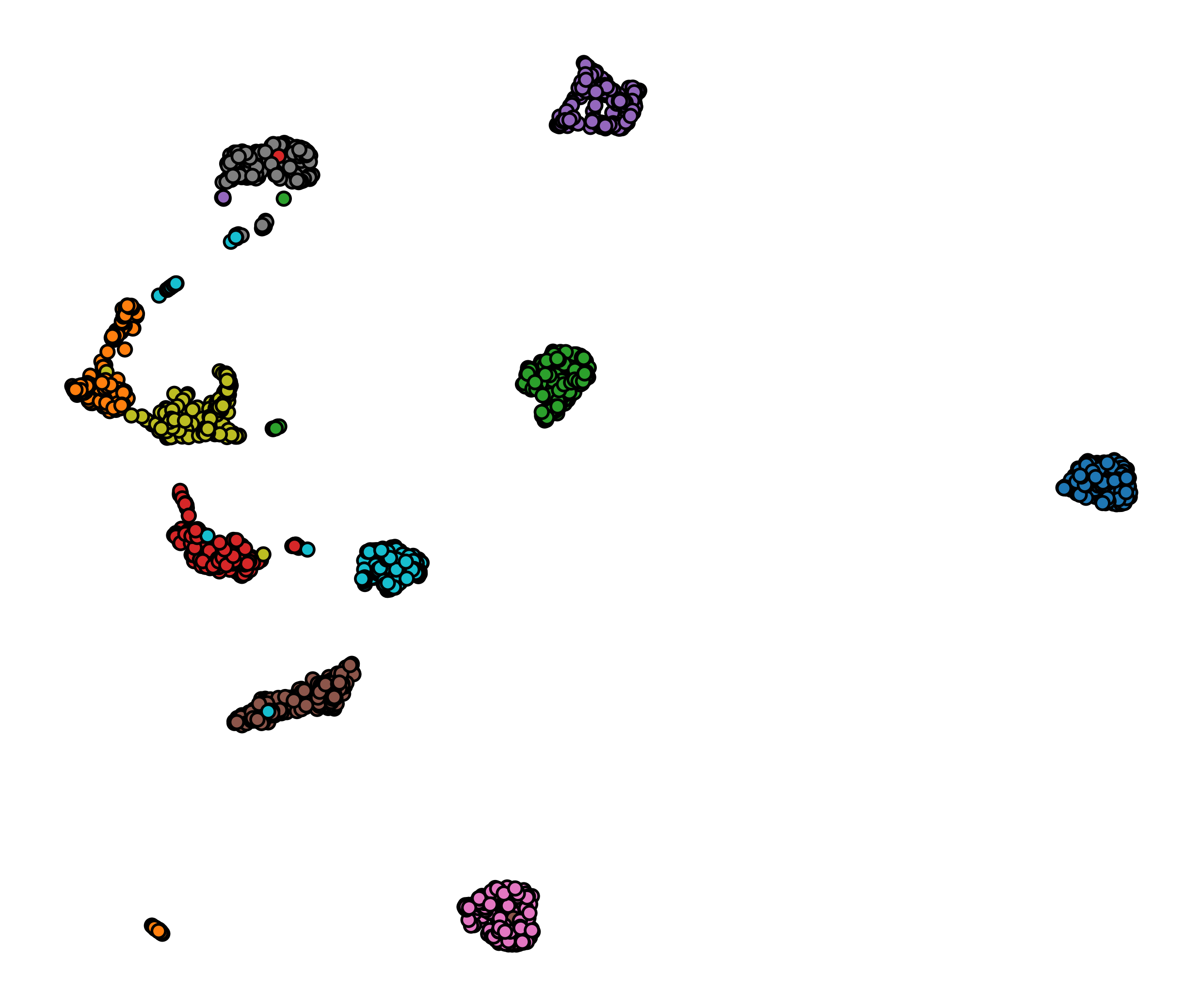}
  \caption{Original UMAP.}
  \label{fig:umap}
\end{subfigure}%
\caption{Comparing different t-SNE and UMAP layouts. UMAP and t-SNE modeled relationships are remarkably similar, and when mapped using the same embedding algorithm resulted in very similar outcomes. However, the original t-SNE and UMAP techniques produce very different results, in terms of groups separation and cohesion, indicating that the mapping phase used by both approach have a considerable impact in the produced layouts.}
\label{fig:local}
\end{figure}

For this specific dataset and employed distance metric, the modeled relationships have a clear importance on the groups' separation. However, the mapping process also impacts that and the final group's density, especially for local DR techniques. Figure~\ref{fig:local} presents examples to illustrate this effect. Figure~\ref{fig:gd_tsne} shows the spring layout embedding of the probability graph generated by t-SNE setting $perplexity=10$. To create this graph, we removed the edges with near-zero probabilities. Figure~\ref{fig:gd_umap} shows the spring layout of the probability graph generated by UMAP setting the \textit{number of neighbors} to 10. To generate these layouts, the spring layout technique was used. The resulting layouts are remarkably similar (also very similar to the NNG graph of Figure~\ref{fig:knn}). The well-separated groups are clearly visible in both layouts and mixed parts are more or less the same, with a few minor differences, e.g., the orange small sub-group that seems to fluctuate around the classes not being well-connected to the other items in the orange class. 

When comparing the t-SNE and UMAP graph layouts with the layouts produced by the original t-SNE and UMAP techniques, Figures~\ref{fig:tsne}~and~\ref{fig:umap}, respectively, the impact of the mapping phase becomes evident. The layouts produced by t-SNE and UMAP are much more clustered and well-separated, with still the same sub-clusters (light blue and orange) being noticeable. Also, the main difference between t-SNE and UMAP is clear. UMAP produces much more compact clusters, mainly due to the larger attraction forces resulting from the employed negative sampling strategy~\cite{JMLR:v23:21-0055}. Indeed, considering our framework, it is possible to say that both techniques model very similar relationships but have very different visual outcomes, mainly due to the mapping phase. Notice that we are not implying that this is the general case, but, in our example, and under the adopted parametrization, it is clear that the different mappings substantially contribute to making both layouts notably different, much more than the modeled relationships.

By considering our framework and the opportunities it opens in adopting graph metrics to validate DR layouts, it is also possible to quantify the differences between the modeled relationships and the impact of the mapping strategies. A simple idea is to use the faithfulness metric (see Section~\ref{sec:qualityanalysis}). The idea of faithfulness is to calculate the Jaccard score, the product of the edges intersection by the edges union, of two graphs with the same nodes but different edge sets. With that, we can, for instance, quantify differences of UMAP and t-SNE modeled relationships. For the graphs of Figures~\ref{fig:gd_tsne}~and~\ref{fig:gd_umap}, the faithfulness is $0.86$, indicating that roughly $86\%$ of the edges are common. When comparing both graphs with the k-NN graph (Figure~\ref{fig:knn}), faithfulness is $0.93$ and $0.86$, respectively, indicating that UMAP is more similar to a KNN graph than a t-SNE graph. We are ignoring edge weights in this computation, but this opens some interesting opportunities when discussing the level of similarity between the relationships modeled by different techniques. 

Faithfulness can also be used, as originally proposed, to evaluate produced layouts in relation to the modeled relationships, when a graph is compared against a proximity (shape) graph calculated on the visual representation. Since the graph modeling the relationships in these examples is a type of NNG, we calculated a k-NNG of the 2D mapping as a proximity graph --  we set $k=10$ to create the 2D k-NNG to match the number of neighbors of the modeled relationships. The faithfulness of the t-SNE mapping is $0.39$ while UMAP attains $0.34$ (Figures~\ref{fig:tsne}~and~\ref{fig:umap}).  For comparison, the spring layouts of t-SNE and UMAP (Figures~\ref{fig:gd_tsne}~and~\ref{fig:gd_umap}) achieved $0.22$ and $0.20$, respectively, supporting that UMAP and t-SNE mapping strategies are better than a simple spring layout and that the t-SNE mapping can create a layout that better represents the modeled relationships. It is also possible to compare layouts by comparing their 2D k-NNGs graphs. For instance, the faithfulness of the original t-SNE and UMAP layouts is $0.45$, indicating that the neighborhood representations conveyed by both layouts are around $50\%$ similar.

\begin{figure}
\begin{subfigure}{.5\linewidth}
  \centering
  \includegraphics[width=\linewidth]{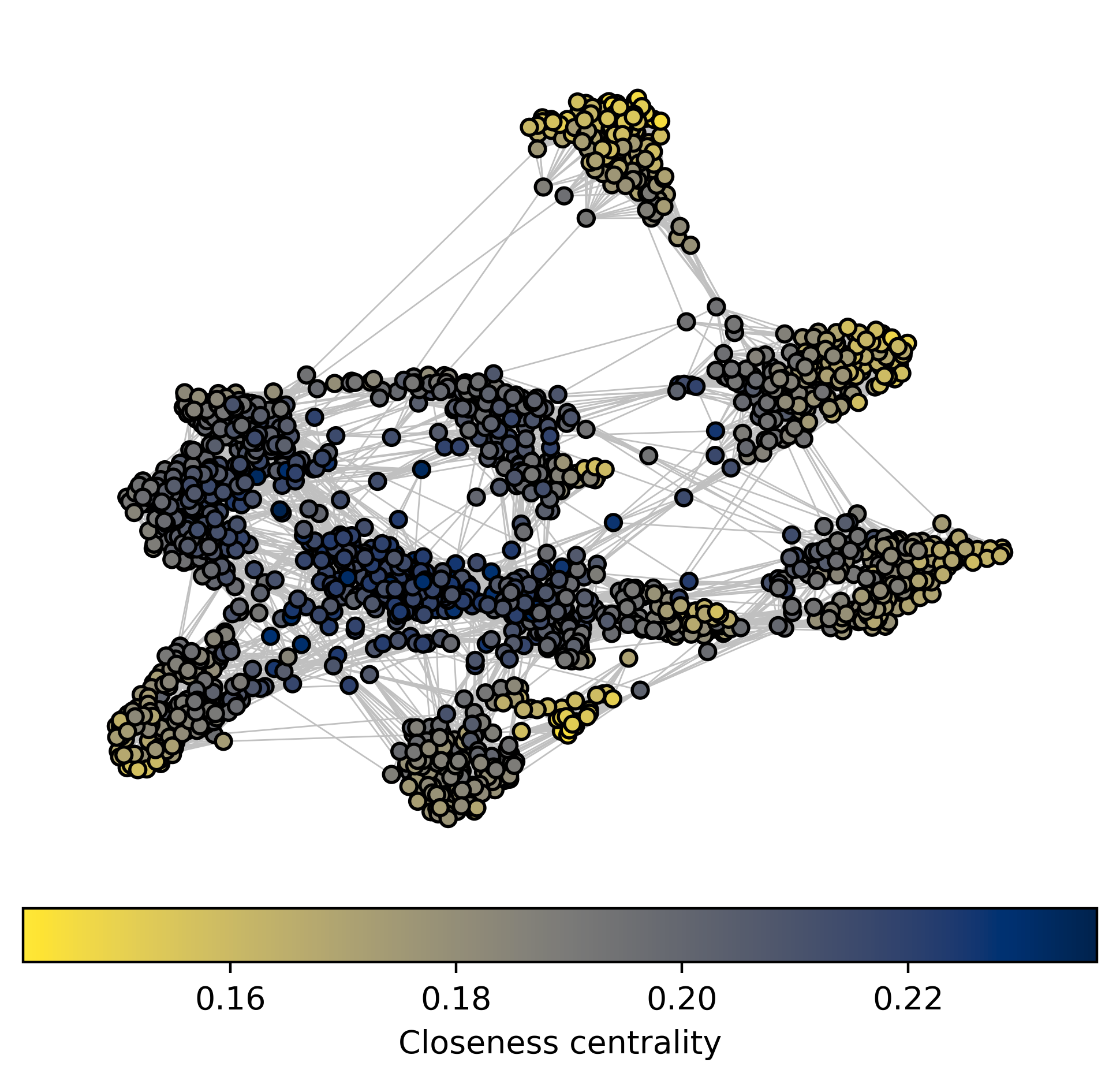}
  \caption{t-SNE graph.}
  \label{fig:gd_tsne_closeness_centrality}
\end{subfigure}
\begin{subfigure}{.5\linewidth}
  \centering
  \includegraphics[width=\linewidth]{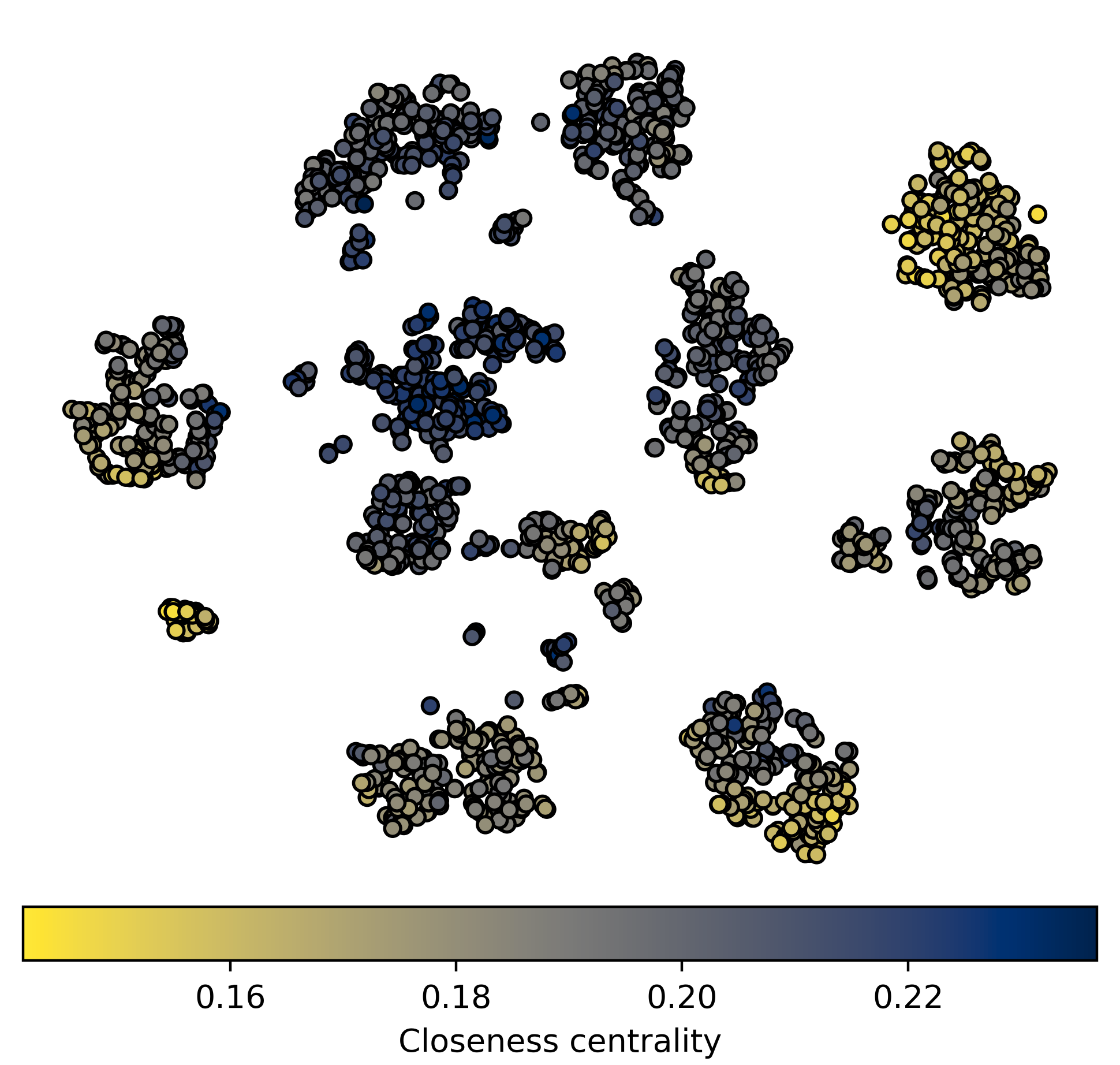}
  \caption{Original t-SNE.}
  \label{fig:tsne_closeness_centrality}
\end{subfigure}\\
\begin{subfigure}{.5\linewidth}
  \centering
  \includegraphics[width=\linewidth]{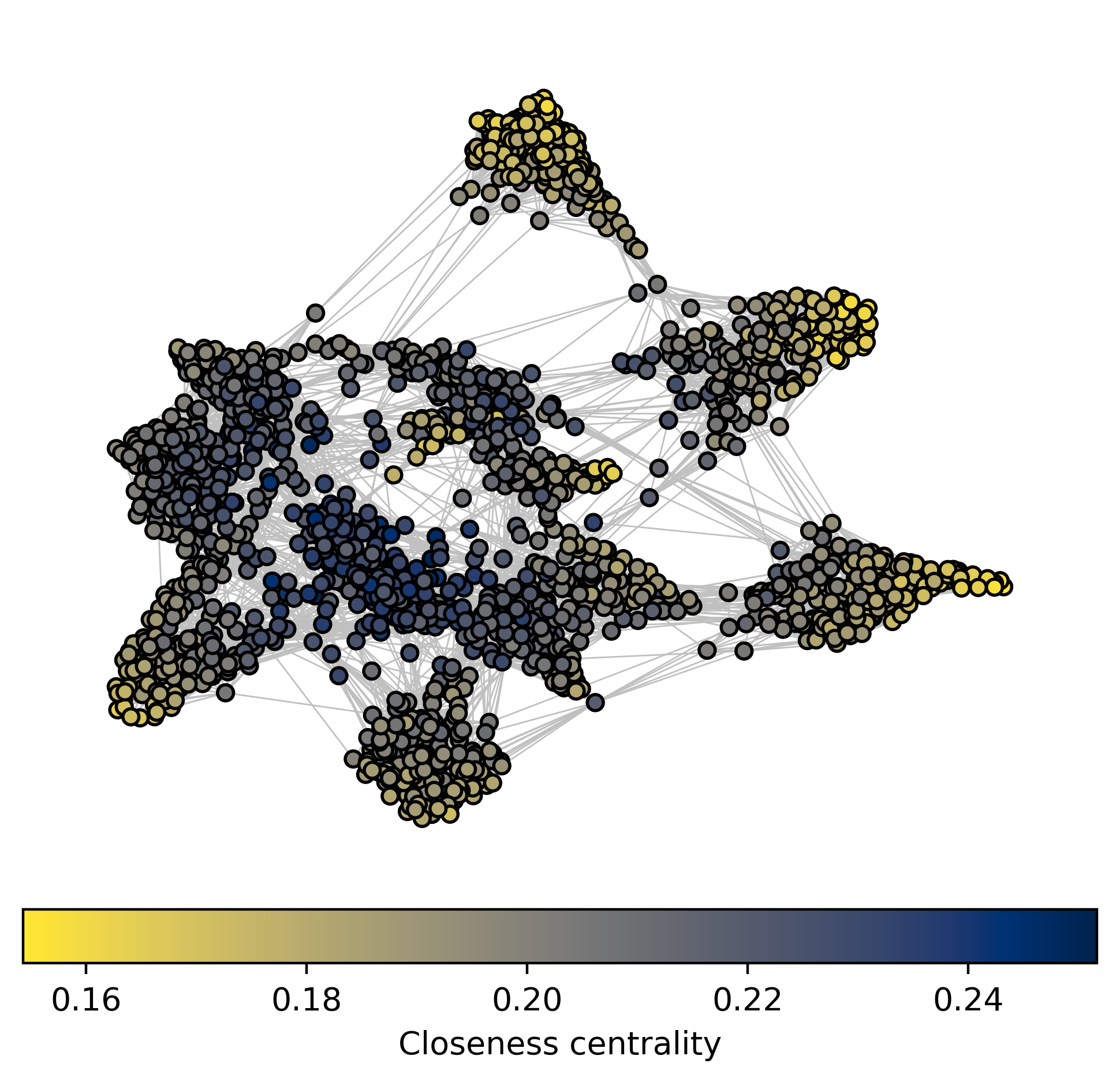}
  \caption{UMAP graph.}
  \label{fig:gd_umap_closeness_centrality}
\end{subfigure}
\begin{subfigure}{.5\linewidth}
  \centering
  \includegraphics[width=\linewidth]{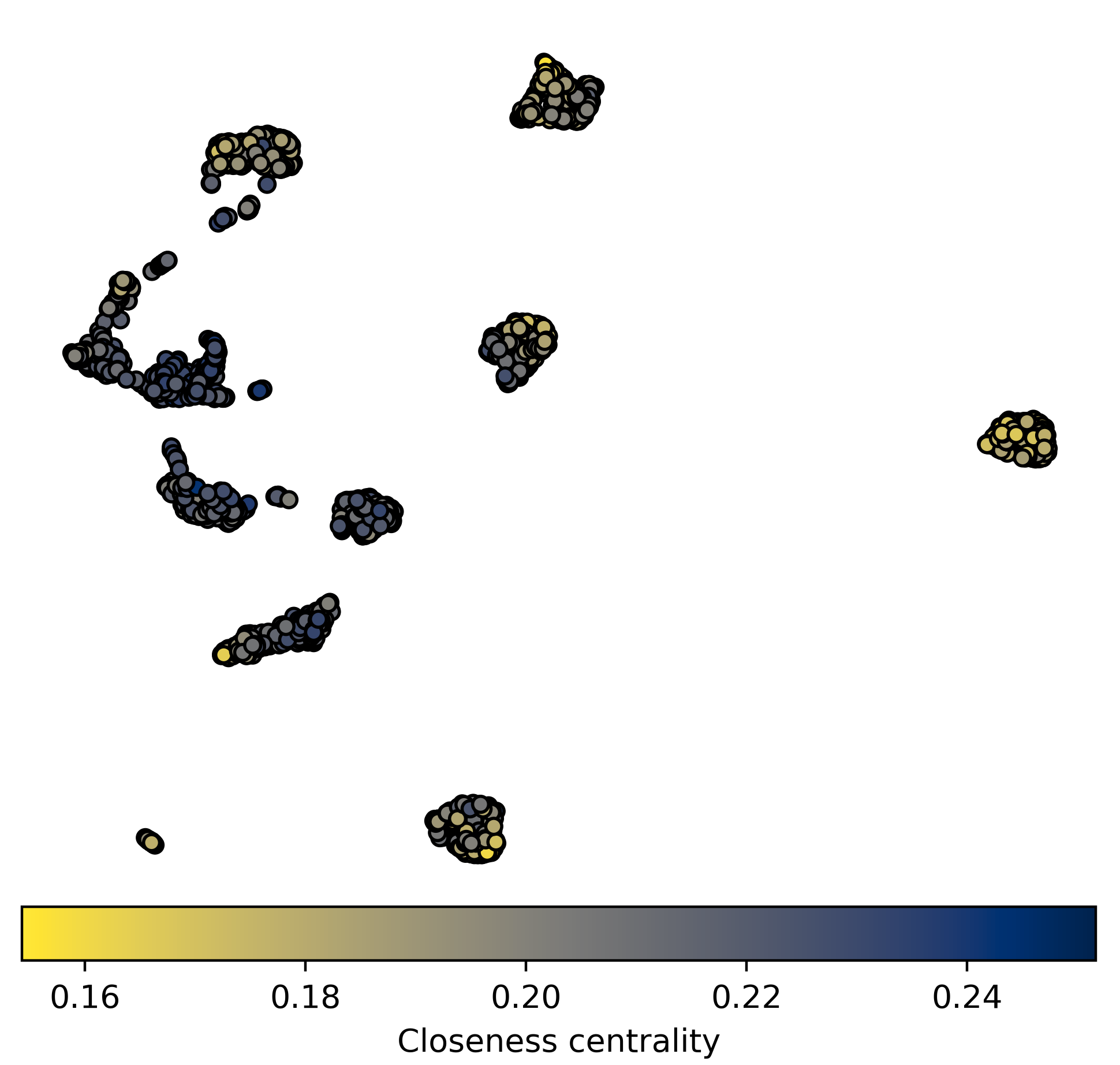}
  \caption{Original UMAP.}
  \label{fig:umap_closeness_centrality}
\end{subfigure}%
\caption{Nodes' closeness centrality of different t-SNE and UMAP layouts. Groups and sub-groups of nodes presenting low closeness indicates well-defined groups, including the orange sub-group which is well separated from the other oranges nodes considering the modeled relationships.}
\label{fig:centrality}
\end{figure}

Metrics are summarizations, so details to help validate individual nodes (data items), groups of nodes, or even individual edges are lost. A more informative way would be to use some graph measure to quantify specific modeling (or graph) properties. One possibility is to use one of the existing centrality measures, for instance, the closeness centrality, or simply closeness~\cite{Sabidussi:Psychometrika66}. Closeness of a node is the sum of the shortest paths between that node and all other nodes, so that larger centrality indicates nodes that are close to all other nodes. Figure~\ref{fig:centrality} shows the UMAP and t-SNE layouts and their respective graphs colored based on their nodes' closeness. Both for t-SNE and UMAP relationships (Figures~\ref{fig:gd_tsne_closeness_centrality}~and~\ref{fig:gd_umap_closeness_centrality}), the well-separated groups (blue, pink, and purple) present somewhat smaller values of closeness, indicating that their nodes are not close to other nodes in the graph, being a distinct group based on the modeled relationships and visually confirmed by the resulting layouts. That also allows us to verify sub-groups, for instance, the small sub-cluster of the orange nodes. It has also very low closeness while the remaining orange nodes have much larger values, so an indication that under the employed parametrization, such a small cluster resulted from the modeling and not an artifact of the mapping stage. 

\section{Discussion, Limitations, and Future Work}
\label{sec:discussion}

In our experiments, we discussed only a few examples of how our framework can be used in practice to better understand the impact of the modeling and mapping phases of DR techniques. We focus on illustrating some possibilities and not comprehensively exploring the best combinations or strategies. So the conclusions and discussions presented in the previous section are limited to their scope. They cannot be generalized outside the \textbf{digits} dataset and the employed parametrization. The idea is just to point out some interesting practical directions for further explorations and not to support that a technique is similar or dissimilar, better or worse than any other technique. We left as future work the possibility of comparatively surveying multiple DR techniques to support more generalized conclusions and shed some light on the real differences in the existing literature.

Although our framework is generic enough to fit a range of different global and local techniques, enabling us to dissect techniques in terms of the modeled relationships and the embedding processes, it cannot accommodate some techniques that do not focus on representing relationships between data items. One important example is Principal Components Analysis (PCA). On PCA, the goal is to rotate a dataset so that its variance can be captured as much as possible with a few axes (or components). Although through mathematical manipulation it is possible to transform such a problem into a pairwise (Euclidean) distance preservation -- PCA produces the same results as the classical MDS technique~\cite{van2009dimensionality} -- the goal is not semantically focused on relationships. This is also true for other linear techniques, such as the Random Projection~\cite{ACHLIOPTAS2003671}, or any technique where the distinction between what is modeled and how it is mapped is not clear, or that admits approximations as the out-of-sample strategies, or uses information of the input dataset during the mapping stage that is not modeled as relationships. So, there is still a portion of the DR literature which cannot be successfully represented to take advantage of what is offered by graph drawing theory.

One of the limitations of our paper is not diving into the possibilities of how graph visualization, interaction and analytics tooling can be adapted to support DR pipelines. Visualization and interaction is a vast field with many ramifications and definitions and a complete study deserves a separate paper. For instance, the non-Euclidean mappings could be explored, like the DoughNet~\cite{bauer2025doughnt}, which maps graphs to a torus, or the mapping to a sphere~\cite{wageningen2024experimental}, allowing to navigate the DR layout to better represent local relationships. Also,  the idea of clicking on a point and attracting similar points that were wrongly mapped far away, similar to what is proposed in~\cite{Tominski2024} for graph nodes, where the connected nodes to a node are attracted, is another option of possible interaction.

The major focus of this paper is to formalize and formulate a framework to map DR and GD so that DR can benefit from unexplored opportunities. Important to notice that some techniques already  consider the opposite perspective by using DR to support GD. Examples are TsNET~\cite{kruiger2017tsnet} and DRGraph~\cite{minfeng2021drgraph}, which use the t-SNE strategy to draw graphs. We refrain from exploring such space and leave the exploration of how DR can attend GD process for future work.


Finally, the claim that DR and GD are somewhat similar has already been pointed out~\cite{mcinnes2018umap, JMLR:v23:21-0055} but never systematically defined or discussed in the light of how GD can support DR pipelines. Therefore, not the concept but the explicit definition and attempt to formalize is our main contribution. Through such systematization, we open different opportunities to enhance the design and validation of DR techniques or pipelines, focusing on the intended relationships to represent and also the goals of the produced visual representations.

\section{Conclusions}
In this paper, we present a framework for bridging DR and GD. We divide the DR visualization pipeline into four stages - that are \textit{Relationships}, \textit{Mapping}, \textit{Quality Analysis}, \textit{Visualization and Interaction}. Specifically, we focus on how GD techniques could be used in the context of DR to improve and contribute to each of these stages. We base and ground our considerations on the similarities between the GD and DR techniques. Future work includes further experiments on the opportunities that we outline in this paper. Our paper does not have the objective of being fully exhaustive, but rather we aim to encourage the discussion and foster further research at the intersection of GD and DR, potentially approximating the DR and GD communities, leveraging opportunities beyond what is discussed in this paper.

\bibliographystyle{eg-alpha-doi}  
\bibliography{sample}        


\end{document}